\documentclass[lettersize,journal]{IEEEtran}
\usepackage{amsmath,amsfonts}
\usepackage{algpseudocode}
\usepackage{array}
\usepackage[caption=false,font=normalsize,labelfont=sf,textfont=sf]{subfig}
\usepackage{textcomp}
\usepackage{stfloats}
\usepackage{url}
\usepackage{verbatim}
\usepackage{graphicx}
\usepackage{booktabs}
\usepackage{tabularx}
\usepackage{multirow}
\usepackage{makecell}
\usepackage{xcolor}
\usepackage{subcaption}
\usepackage{algorithm}
\usepackage{graphicx}
\usepackage{placeins}
\def\BibTeX{{\rm B\kern-.05em{\sc i\kern-.025em b}\kern-.08em
    T\kern-.1667em\lower.7ex\hbox{E}\kern-.125emX}}
\usepackage{balance}
\usepackage[style=ieee]{biblatex}
\AtEveryBibitem{
  \clearfield{issn}
  \clearfield{isbn}
  \clearfield{pmid}
  \clearfield{doi}
}

\begin{document}

\title{Neural Ordinary Differential Equations for Simulating Metabolic Pathway Dynamics from Time-Series Multiomics Data}

\author{Udesh Habaraduwa* and Andrei Lixandru*
    \thanks{*U. Habaraduwa and A. Lixandru contributed equally to this work.}%
    \thanks{U. Habaraduwa is with Amsterdam University Medical Center.}%
    \thanks{A. Lixandru is with Griffin Labs.}%
}

\maketitle

\vspace{1em} 

\begin{abstract}
The advancement of human healthspan and bio-engineering relies heavily on predicting the behavior of complex biological systems. While high-throughput multiomics data is becoming increasingly abundant, converting this data into actionable predictive models remains a bottleneck. High-capacity, data-driven simulation systems are critical in this landscape; unlike classical mechanistic models restricted by prior knowledge, these architectures can infer latent interactions directly from observational data, allowing for the simulation of temporal trajectories and the anticipation of downstream intervention effects in personalized medicine and synthetic biology. To address this challenge, we introduce Neural Ordinary Differential Equations (NODEs) as a dynamic framework for learning the complex interplay between the proteome and metabolome. We applied this framework to time-series data derived from engineered \textit{Escherichia coli} strains, modeling the continuous dynamics of metabolic pathways. The proposed NODE architecture demonstrates superior performance in capturing system dynamics compared to traditional machine learning pipelines. Our results show a greater than 90\% improvement in root mean squared error over baselines across both Limonene (up to 94.38\% improvement) and Isopentenol (up to 97.65\% improvement) pathway datasets. Furthermore, the NODE models demonstrated a $1000\times$ acceleration in inference time, establishing them as a scalable, high-fidelity tool for the next generation of metabolic engineering and biological discovery.
\end{abstract}

\section{Introduction}
For most of history human life span was limited and one would be considered lucky to reach the age of 35. As our understanding of biological processes gradually improved (e.g., the germ theory of disease, antibiotics, antiseptics, etc.), so too the length and quality of life around the world \cite{berg2018longevity}. Today, the expected lifespan of an adult is nearly twice that of 200 years ago \cite{berg2018longevity}. The bottlenecks that stand in the way of yet another doubling may be qualitatively different in nature, necessitating a deeper understanding of the complex dynamical systems at the heart of biological processes \cite{ideker2006bioengineering}. 

Modern bio-engineering techniques (e.g., genomic engineering with CRISPR \cite{ran2013genome} and activation of limb regrowth in species without the innate capacity to do so \cite{murugan2022acute}) have opened up promising avenues for altering the ``natural'' course of biology to achieve more desirable outcomes. However, there is progress yet to be made in predicting the behavior of biological systems as a whole when a change is introduced \cite{costello2018machine}. 

Most systems in biology are observed as short, noisy, time-series recordings (e.g., concentrations of a metabolites or proteins over time) that can be time consuming and expensive to measure \cite{berg2018longevity}. However recent advances in high-throughput techniques (e.g., at the genome, protieome, transcriptome, and metabolome levels) \cite{juan2023quantitative} have resulted in a drastic increase in usable data. Thus, the application of modern machine learning (ML) pipelines (e.g., automated model selection and hyper-parameter tuning \cite{olson2016tpot}) is becoming increasingly feasible \cite{mowbray2021machine}.

Within this data-rich landscape, computational simulations (in silico modeling) offer a potent tool for dissecting biological complexity, having previously demonstrated significant utility in drug target identification and mechanism elucidation \cite{target_id_citation}. To fully leverage modern high-throughput data, there is a compelling need for high-capacity, dynamic, data-driven simulation systems. Dynamic modeling is particularly advantageous as it allows for the simulation of temporal trajectories under various perturbation scenarios. This is critical for anticipating the downstream effects of interventions, such as detecting unwanted side effects that may only unravel over time.

Furthermore, a data-driven approach is essential given our frequently incomplete mechanistic understanding of systemic biology. Unlike classical mechanistic models restricted by prior knowledge, data-driven systems can infer latent interactions directly from observational data. This capability leads to simulations that not only match the biological process with higher fidelity but also aid in hypothesis generation for future mechanistic research. Additionally, such adaptability facilitates personalized medicine, allowing simulations to be tailored to individual patient profiles. Finally, to capture the governing dynamics of these systems, modeling architectures must possess high representational capacity. This is necessary to approximate the complex functions driving biological dynamics and to represent key latent variables—factors that fall outside current measuring capabilities yet remain instrumental in driving system behavior.

To this end, drawing inspiration from work done at the transcriptome level \cite{erbe2023transcriptomic}, we introduce Neural Ordinary Differential Equations (NODEs) as a promising dynamic, data-driven, and high-capacity approach. We apply this framework to learn the complex interplay between the proteome (proteins that are created and modified by an organism \cite{dupree2020critical}) and the metabolome (the collection of metabolites responsible for life \cite{peng2015functional}) directly from time-series data.

\section{Materials and methods}
\subsection{Data}

\begin{figure}[ht]
  \centering
  \subfloat[]{\includegraphics[width=0.6\columnwidth]{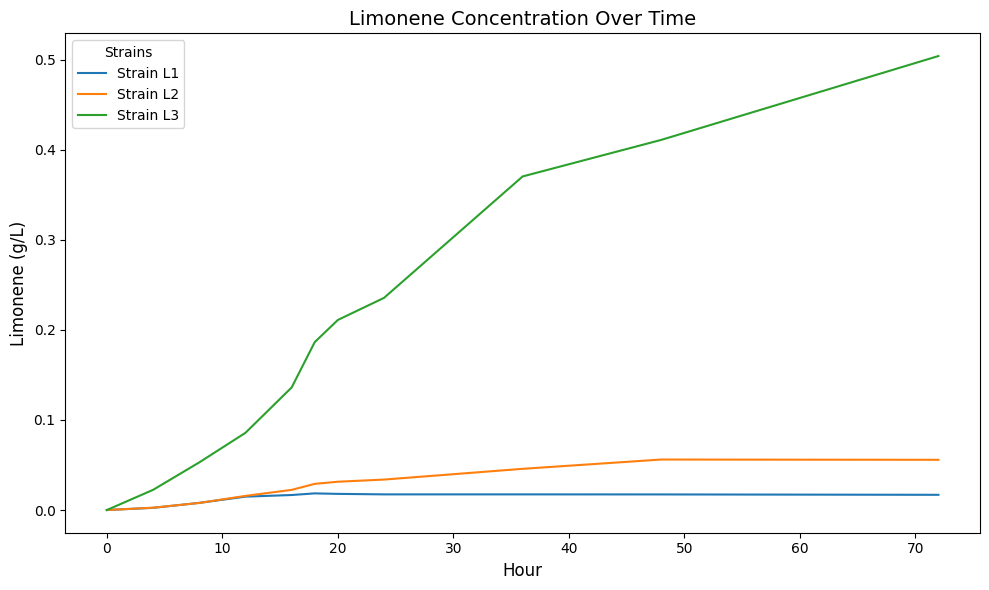}%
  \label{fig:lim_raw}}
  \vspace{-3mm} 
  \\
  \subfloat[]{\includegraphics[width=0.6\columnwidth]{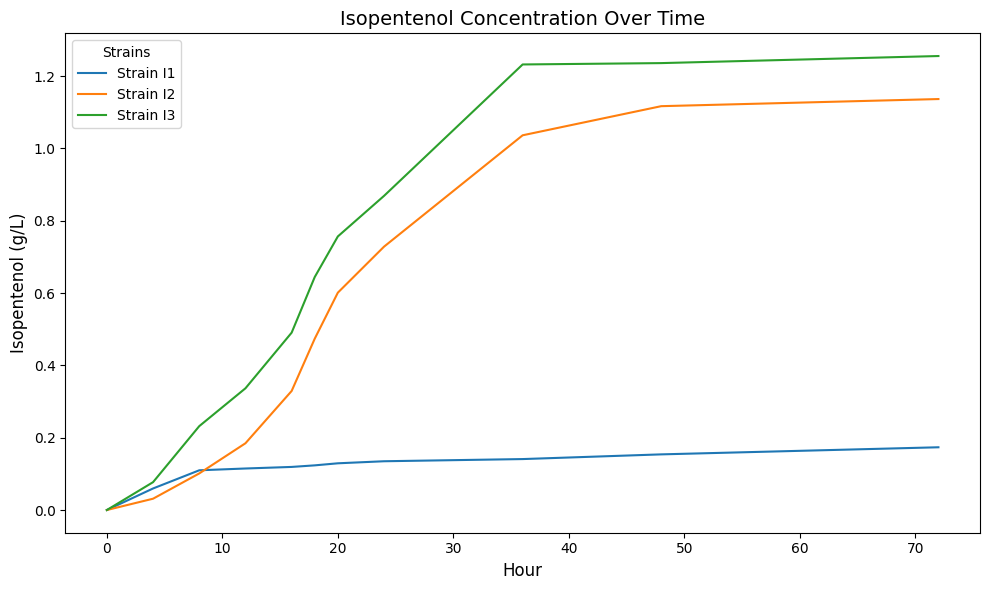}%
  \label{fig:iso_raw}}
  \caption{Time series measurements for metabolite concentrations. (a) Limonene concentrations across strains L1, L2, and L3. (b) Isopentenol concentrations across strains I1, I2, and I3.}
  \label{fig:metabolite_time_series}
\end{figure}

The present study aims to investigate if NODEs can outperform traditional machine learning methods \cite{costello2018machine} in capturing the dynamics of metabolic pathways. To that end, a proteomic and metabolimic time-series dataset is utilized \cite{costello2018machine}. The dataset was derived from an experiment in metabolic engineering \cite{brunk2016characterizing}. Here, different strains of the Escherichia coli bacterium were engineered to be low, medium, or high producers of isopentenol or limonene. The measured time-series for these two metabolites is shown in Figure \ref{fig:metabolite_time_series}.

While limonene and isopentenol are the metabolites of interest, the original dataset contains 91 metabolomic and 23 proteomic features, each recorded over 72 hours in 14 intervals. However, \cite{costello2018machine} only use a subset of these features (Table \ref{tab:controls_states}). The original time-series was interpolated to generate a time-series of 200 intervals. Finally, the two datasets are combined as follows:

\begin{itemize}
    \item Controls - These represent the concentrations of proteins (e.g., AtoB, GPPS, etc.) that regulate the production of metabolites at a given time point. The dataset contains one time-series of concentrations for each protein. Proteins play an important regulatory role within biological systems \cite{shi2024application}. Depending on the protein and the context, they maybe involved in a range of functions from determining cellular structure to facilitating transport \cite{alberts1998cell}. 
    \item States - These represent the concentration of a metabolite of interest (e.g., limonene or isopentenol). Metabolites are the result of metabolic processes and usually the end product of protein metabolism \cite{shi2024application}.  The dataset contains one time-series of concentrations for each metabolite.
\end{itemize}

The controls and states from the data are listed in Table \ref{tab:controls_states}. Each metabolite and protein is represented in a time-series similar to examples shown in Figure \ref{fig:metabolite_time_series}, not shown here for brevity.

\begin{table}[h!]
\centering
\begin{tabular}{p{3cm} p{3cm}}
\hline
\textbf{Controls} & \textbf{States} \\ \hline
AtoB              & Acetyl-CoA      \\ 
GPPS              & HMG-CoA         \\ 
HMGR              & Mevalonate      \\ 
HMGS              & Mev-P           \\ 
Idi               & IPP/DMAPP       \\ 
Limonene Synthase & Limonene         \\ 
MK                & OD600           \\ 
NudB              & GPP             \\ 
PMD               & NAD             \\ 
PMK               & NADP            \\ 
                  & Acetate         \\ 
                  & Pyruvate        \\ 
                  & citrate         \\ 
                  & Isopentenol     \\ 
\hline
\end{tabular}
\caption{List of Controls and States relevant to the limonene and isopentenol pathways, for a total of 23 features with 600 rows (200 for each strain). The limonene and isopentenol pathways were trained independently.}
\label{tab:controls_states}
\end{table}

The dataset can be found by referring to the supplementary material of \cite{costello2018machine}. 

\subsection{Methods}
To begin, let us define the relationship between metabolites and proteins as a differential equation:
\begin{equation}
\dot{m} = f(p, m)
\label{eq:metabolite_protein_relationship}
\end{equation}

where $m$ and $p$ represent metabolite and protein concentrations respectively. This equation states that the change in $m$ and $p$ is a function of the current concentration of $m$ and $p$. Such equations are the preferred mathematical representation of bio-chemical kinetics \cite{costello2018machine,de2024physiology}. 

The models trained by \cite{costello2018machine} serve as a baseline for testing the proposed implementation. In short, their approach is as follows:
\begin{enumerate}
    \item For each state and control variable, approximate the derivatives at each time point. The noisy experimental data is first smoothed and the derivative is calculated using a central differencing scheme \cite{costello2018machine}.
    \item Find a function that best predicts the approximated derivatives. The authors employ an automatic machine learning approach (with TPOT \cite{olson2016tpot}), searching over a space of machine learning models (e.g., Gradient boosting regressor, random forest regressor, etc.) to find the best one. The models are trained on all the states and controls to minimize the loss on predicting the derivatives of the states alone. A unique model is trained for each state derivative. That is, each model takes the place of the right hand side of equation \ref{eq:metabolite_protein_relationship} for one of the metabolites.
    \item The trained models, which serves as a derivative function for each state variable, are then integrated using a ODE solver to generate new state values at a given time point. 
\end{enumerate}

Using a NODE, the step of approximating derivatives and subsequently fitting an ML model for each feature can be circumvented entirely. Instead, a fully connected neural network is trained to learn the derivative of each state and control variable directly. This is represented in equation \ref{eq:neural}, where $f$ is a function of the neural network parameters $\theta$ and the state variables $m$ and $p$ (collectively defined as $u$). Once the network has been trained, it serves as the derivative function in the ODE solve (algorithm \ref{alg:optimization}). 

\begin{equation}
\dot{u} = f(u,\theta)
\label{eq:neural}
\end{equation}

\begin{algorithm}
\caption{Optimization Process}
\label{alg:optimization}
\begin{algorithmic}[1]
\For{\textbf{each} optimization step}
    \State Solve $ \frac{du}{dt} = f(\theta, u) $ to get a trajectory for each feature by integrating the NODE.
    \State Calculate 
    \State Update $ \theta $
\EndFor
\end{algorithmic}
\end{algorithm}

Each solve step of algorithm \ref{alg:optimization} results in a vector of shape $ 200 \times 23 $. Consider one row of the dataset ($ 1 \times 23$), representing one time point for each feature. The solve step returns a trajectory matrix of shape $200 \times 23$ : The data point evolved for 200 time steps. The last time step from each trajectory is taken to be the NODE prediction for this time point. That is, the neural network is tasked with learning the vector field that evolves a given time point through a trajectory that best approximates the dynamics represented in the training data \cite{chen2018neural}. 

\begin{figure}[htbp]
    \centering
    \includegraphics[width=0.3\textwidth]{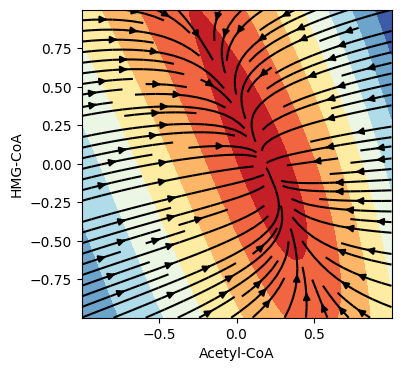}
    \caption{Visualization of a (toy) learned vector field in two possible dimensions. Color intensity represents the magnitude of the vector field, with red indicating regions of faster dynamics and blue showing slower changes. Black streamlines indicate the direction of system evolution.}
    \label{fig:vectorfield}
\end{figure}

For illustration, consider the toy space defined by two features (e.g., Acetyl-CoA and HMG-CoA) in Figure \ref{fig:vectorfield}. The model is trained to learn the ``shape'' of this space such that for a given training point ( e.g., $[\text{HMG-COA}[t_0],\text{Acetyl-CoA}[t_0]]$), the point evolves over $n$ time-steps and ends up in the ``right place''.

A promising approach to training NNs in physical applications is to incorporate domain knowledge into the model \cite{de2024physiology}. Here, a physiology-informed regularization term (PIR) ( $\lambda = [0.01,1.0,1000])$ is added to the loss function which penalizes predictions of negative values for metabolite concentrations (equation \ref{eq:neg_penalty}).  Optimization is carried out in two stages \cite{de2024physiology}. First, with ADAM \cite{kingma2014adam} for 300 epochs ($ \text{learning rate (LR)} = 1e-4$) and BFGS \cite{fletcher2000practical}($\text{LR} = 1e-4, \text{maximum iterations} = 1000, \text{gradient tolerance} = 1e-6,\text{change tolerance} = 1e-6$).

\begin{equation}
    \mathcal{L}_{\text{total}} = \mathcal{L}_{\text{MSE}} + \lambda \cdot \frac{1}{n}\sum_{i=1}^{n} (\min(0, \hat{y_i}))^2
    \label{eq:neg_penalty}
\end{equation}

The architecture of the NN component of the NODE is shown in Table \ref{tab:neuralode}. In terms of architectural decisions, considerations were made regarding the size (depth and width) and activation functions. Though it was assumed that general characteristics of NNs (e.g., depth improves performance \cite{safran2017depth}) would hold, a relatively small NN was chosen (useful in scientific machine learning applications \cite{rackauckas2020universal} where the outputs of a NN can be directly linked to, for example, unobserved interactions of metabolites). The chosen activation function should be differentiable and smooth \cite{anonymous2024exploring, kidger2022neuraldifferentialequations}, both of which the hyperbolic tangent function satisfy \cite{kidger2022neuraldifferentialequations}. 

A prediction of a trajectory requires an ODE solve to approximate the current value of the function (i.e., the approximation of the current concentration of a metabolite given its derivative). A variety of methods exist for obtaining approximate solutions to time-dependent ordinary differential equations (e.g., the Runge-Kutta family of methods \cite{dormand1980family}). Here, the 8th order Dormand-Prince \cite{dormand1980family} method is used from the torchdiffeq library \cite{torchdiffeq} with default error tolerance parameters.

\begin{table}[h!]
\centering
\caption{Architecture of the NeuralODE Function. Weights are initialized with mean 0 and standard deviation 0.1.}
\label{tab:neuralode}
\begin{tabular}{@{}ccc@{}}
\toprule
\textbf{Layer} & \textbf{Type}               & \textbf{Details}                       \\ \midrule
1              & Linear                     & Input: 23, Output: 10, Bias: True      \\
2              & Activation (Tanh)          & -                                      \\
3              & Linear                     & Input: 10, Output: 10, Bias: True      \\
4              & Activation (Tanh)          & -                                      \\
5              & Linear                     & Input: 10, Output: 10, Bias: True      \\
6              & Activation (Tanh)          & -                                      \\
7              & Linear                     & Input: 10, Output: 10, Bias: True      \\
8              & Activation (Tanh)          & -                                      \\
9              & Linear                     & Input: 10, Output: 23, Bias: True      \\ \bottomrule
\end{tabular}
\end{table}

Code is available online \cite{habaraduwa2024}. All experiments were conducted on a Google Colab instance with the following specifications:
\begin{itemize}
    \item CPU: Intel(R) Xeon(R) @ 2.00GHz (8 cores)
    \item GPU: NVIDIA Tesla T4 (15.36GB VRAM)
    \item Operating System: Linux-6.1.85+-x86\_64
    \item Python Version: 3.10.12
    \item PyTorch : 2.5.1+cu121
    \item torchdiffeq : 0.2.5
    \item Numpy : 1.26.4
\end{itemize}

\section{Results}

\begin{figure*}[t]
  \centering
  \subfloat[]{\includegraphics[width=0.32\textwidth]{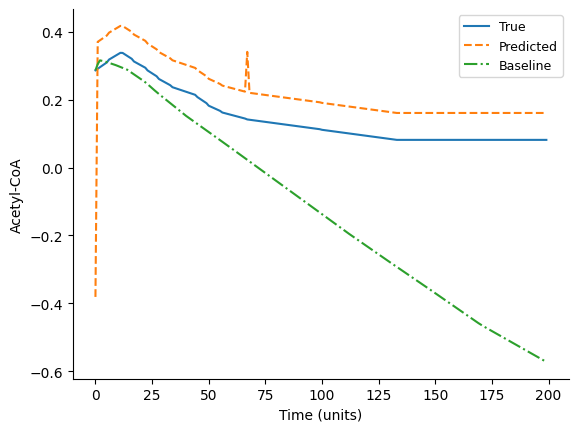}%
  \label{fig:lim-acetyl-coa}}
  \hfill
  \subfloat[]{\includegraphics[width=0.32\textwidth]{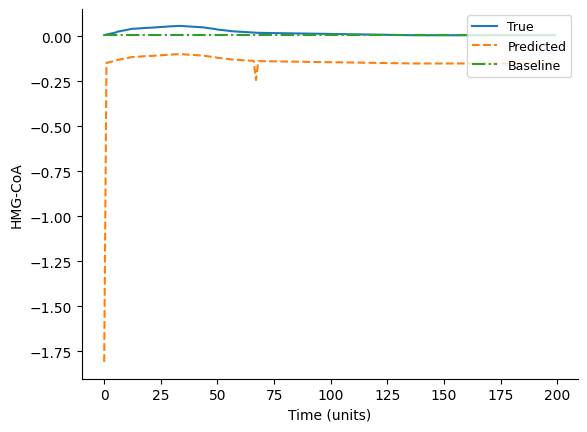}%
  \label{fig:lim-hmg-coa}}
  \hfill
  \subfloat[]{\includegraphics[width=0.32\textwidth]{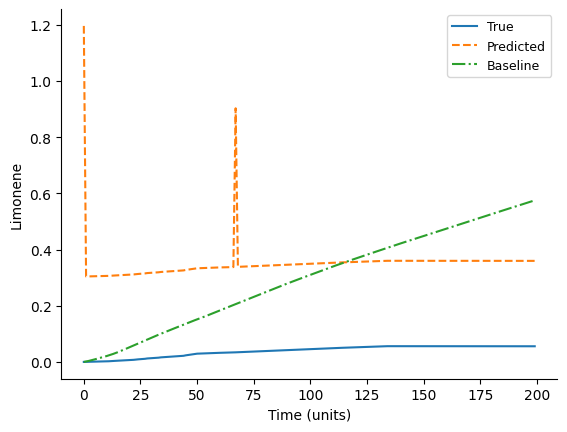}%
  \label{fig:lim-limonene}}
  \\
  \subfloat[]{\includegraphics[width=0.32\textwidth]{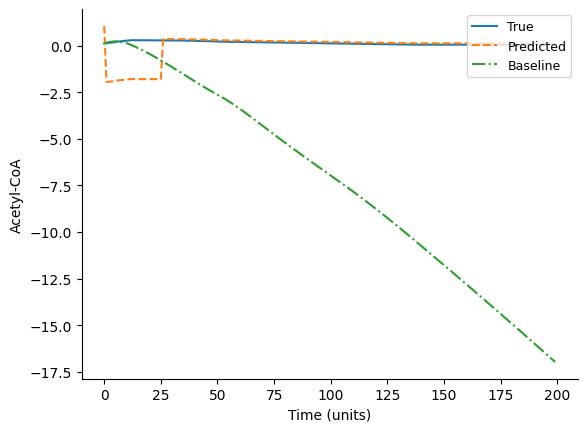}%
  \label{fig:iso-acetyl-coa}}
  \hfill
  \subfloat[]{\includegraphics[width=0.32\textwidth]{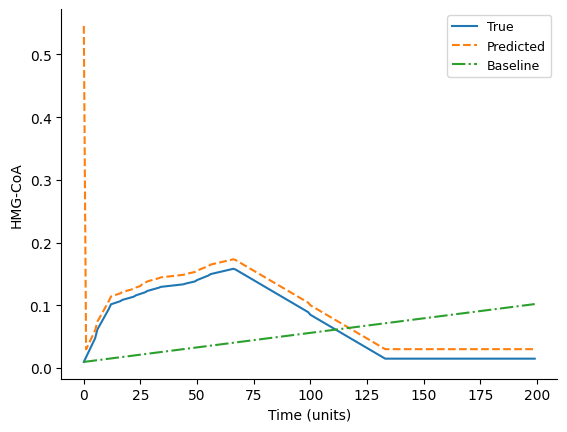}%
  \label{fig:iso-hmg-coa}}
  \hfill
  \subfloat[]{\includegraphics[width=0.32\textwidth]{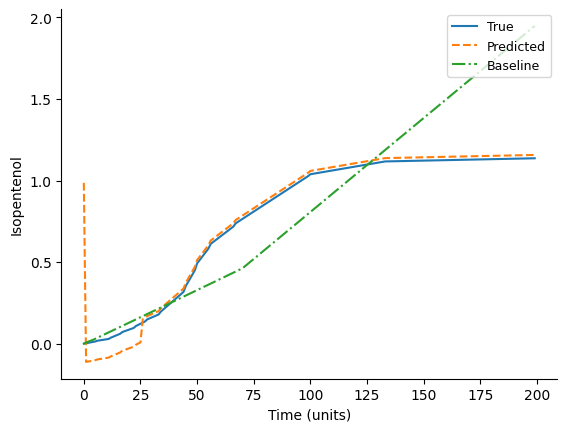}%
  \label{fig:iso-isopentenol}}
  \caption{Prediction of strain dynamics for Limonene and Isopentenol pathways on the held out medium producing strains (I2 and L2). 
  Limonene pathway (top row): (a) Acetyl-CoA concentration, (b) HMG-CoA concentration, (c) Limonene concentration. 
  Isopentenol pathway (bottom row): (d) Acetyl-CoA concentration, (e) HMG-CoA concentration, (f) Isopentenol concentration. 
  All experimental data was recorded over 72 hours, every 2 hours. Data was interpolated to 200 time points\cite{costello2018machine}. Only a subset of all predicted metabolites shown. Note that the NODE is able to qualitatively capture the underlying pattern even if it is off by a scale factor.}
  \label{fig:pathway-dynamics}
\end{figure*}

The performance compared to the baseline models of \cite{costello2018machine} are shown in Tables \ref{tab:perf_lim} and \ref{tab:perf_iso}. Plots of the trajectories generated by the best performing model for Limonene strain dynamics is shown in Figure \ref{fig:pathway-dynamics}. All models were trained on low and high producing strain data and tested on the held out medium producing strain (as in \cite{costello2018machine}).

\begin{table}[H]
\centering
\caption{Performance of NODE over baseline on the Limonene test data for varying regularization strengths ($\lambda$). Values are root mean squared error (RMSE).}
\label{tab:perf_lim}
\begin{tabular}{lrrrr}
\hline
            & \textbf{0.01} & \textbf{1.00} & \textbf{1000.00} & \textbf{Baseline} \\ \hline
Acetyl-CoA  & 0.20          & 0.09          & 1.07             & 0.34              \\
HMG-CoA     & 0.50          & 0.20          & 0.53             & 0.02              \\
Mevalonate  & 1.21          & 0.37          & 2.38             & 1.07              \\
Mev-P       & 1.30          & 0.17          & 0.23             & 3.64              \\
IPP/DMAPP   & 0.36          & 0.37          & 0.32             & 75.43             \\
Limonene    & 0.57          & 0.32          & 0.15             & 0.30              \\
OD600       & 0.91          & 0.61          & 2.25             & 4.30              \\
GPP         & 1.02          & 0.54          & 0.61             & 0.07              \\
NAD         & 0.63          & 0.35          & 0.16             & 1.50              \\
NADP        & 0.07          & 0.50          & 1.22             & 1.26              \\
Acetate     & 0.44          & 0.60          & 1.20             & 1.89              \\
Pyruvate    & 0.33          & 0.15          & 0.44             & 0.14              \\
Citrate     & 0.28          & 0.83          & 0.70             & 0.23              \\ \hline
\textbf{Mean RMSE}          & \textbf{0.60} & \textbf{0.39}    & \textbf{0.87}    & \textbf{6.94}    \\
\textbf{\% Improvement} & \textbf{91.35}  & \textbf{94.38}   & \textbf{87.46}   & \textbf{--}    \\ \hline
\end{tabular}
\end{table}

\begin{table}[H]
\centering
\caption{ Performance of NODE over baseline on the Isopentenol test data for varying regularization strengths ($\lambda$). Values are RMSE.}
\label{tab:perf_iso}
\begin{tabular}{lrrrr}
\hline
            & \textbf{0.01} & \textbf{1.00} & \textbf{1000.00} & \textbf{Baseline} \\ \hline
Acetyl-CoA  & 0.56          & 0.74          & 0.16             & 9.02              \\
HMG-CoA     & 0.70          & 0.04          & 0.34             & 0.08              \\
Mevalonate  & 1.03          & 0.25          & 0.63             & 2.55              \\
Mev-P       & 0.23          & 0.20          & 0.64             & 126.19            \\
IPP/DMAPP   & 0.22          & 0.23          & 0.65             & 33.77             \\
OD600       & 0.07          & 0.31          & 0.84             & 2.13              \\
GPP         & 0.03          & 0.27          & 1.24             & 0.00              \\
NAD         & 0.29          & 0.61          & 0.72             & 1.96              \\
NADP        & 1.03          & 0.16          & 0.74             & 0.24              \\
Acetate     & 0.25          & 0.69          & 0.10             & 0.45              \\
Pyruvate    & 0.85          & 0.25          & 0.56             & 0.06              \\
Citrate     & 0.13          & 0.37          & 0.33             & 0.29              \\
Isopentenol & 0.10          & 0.08          & 0.25             & 0.32              \\ \hline
\textbf{Mean RMSE} & \textbf{0.42} & \textbf{0.32} & \textbf{0.55} & \textbf{13.62} \\ 
\textbf{\% Improvement} & \textbf{96.92} & \textbf{97.65} & \textbf{95.96} & \textbf{--} \\ \hline
\end{tabular}
\end{table}

\begin{table}[H]
    \centering
    \caption{Training and inference times for baseline and NODE models (averaged over $\lambda$ values). Times may vary slightly between training runs, particularly for baseline models due to TPOT's genetic recombination process. However, the times for both models remained consistent and comparable across runs and datasets (see \cite{habaraduwa2024}).}
    \label{tab:training_inference_times}
    \begin{tabular}{@{}lcccc@{}}
        \toprule
        Model     & \multicolumn{2}{c}{Training Time (minutes)} & \multicolumn{2}{c}{Inference Time (minutes)} \\ 
        \cmidrule(lr){2-3} \cmidrule(lr){4-5}
                  & Limonene & Isopentenol & Limonene & Isopentenol \\ \midrule
        Baseline  & 52.08    &  61.79      & 4.83     & 4.04        \\
        NODE      & 5.61     & 5.40        & 0.004    & 0.004        \\ \bottomrule
    \end{tabular}
\end{table}

\FloatBarrier

\section{Discussion and concluding remarks}
The present study evaluated the performance of NODEs against previous generations of ML techniques \cite{costello2018machine} in learning the dynamics represented in metabolomic and proteomic time-series data.

Across both dynamics datasets (limonene and isopentenol), the NODEs out perform the baseline models with 87.5\% - 95\% improvement in mean RMSE. The models also seem to benefit from a moderate amount of PIR ($\lambda = 1.0$) in learning the dynamics in both datasets. Notably, even when the error between the NODE prediction and actual time series is large, the model appears to qualitatively capture the underlying dynamics, missing only the scale factor (visible in \ref{fig:lim-acetyl-coa}), as in \cite{costello2018machine}. Indeed, for metabolic engineering purposes the ability to reliably identify the most productive strain is sufficient \cite{costello2018machine}. 

The NODE models were approximately 10x faster in training and 1000x faster at inference (Table \ref{tab:training_inference_times}). The long training times taken for the baseline models can be attributed to the process of searching through a space of best pipelines with models that do not inherently support hardware acceleration \cite{scikit-learn-faq}. Nonetheless, even if more modern implementation were to be used, fitting a unique function to serve as the derivative for each state variable may ultimately be unnecessary.

NODEs, much like other ML methods, rely heavily on the quantity and quality of the data, necessitating the interpolation techniques implemented by \cite{costello2018machine}. It's possible that longer and higher resolution time-series data would enable the use of deeper networks for capturing richer representations. There is also room for improvement in the NODE set up used here in. Stiff (neural) ODEs (i.e., where the time-steps necessary for good approximations of the integrated solutions grows extremely small), pose a problem for the computational efficiency which may be addressed by stabilizing gradient calculations, scaling of network outputs, and using alternative activation functions ( or learning the best one for the task and data at hand \cite{wang2023learning,cui2023robustness}). Performance may benefit further from NN specific regularization techniques (e.g., enforcing Lipschitz continuity \cite{gouk2021regularisation}) and incorporating further knowledge of the task domain into the loss function. Finally, given the simplicity of the underlying NN, it may be possible to derive governing equations for the system \cite{forootani2023robustsindyapproachcombining}.  

The results herein illustrate the potential of NODEs for efficiently and accurately capturing the dynamics of interacting metabolic and proteomic systems. For the future of bioengineering, these techniques may prove to be invaluable for rapidly iterating through design space,  control mechanisms in production (e.g., for moving from the lab bench to industrial scale \cite{chubukov2016synthetic}), and generating new insight into the nuanced mechanics of biochemical interaction processes. With better prediction and control of biology, it may be possible to unlock the next stages of longer and healthier lives.

\printbibliography


\end{document}